\documentclass[letterpaper]{article} 
\usepackage{aaai2026}  
\nocopyright 
\usepackage{times}  
\usepackage{helvet}  
\usepackage{courier}  
\usepackage[hyphens]{url}  
\usepackage{graphicx} 
\usepackage{natbib}  
\usepackage{caption} 
\usepackage{microtype}
\usepackage{algorithm}
\usepackage{algorithmic}
\usepackage{xspace}
\usepackage{amsmath}
\usepackage{amssymb}
\usepackage{pifont}
\usepackage{booktabs}
\usepackage{multirow}
\usepackage{array}
\DeclareMathOperator{\BCE}{BCE}
\usepackage{newfloat}
\usepackage{listings}
\DeclareCaptionStyle{ruled}
{labelfont=normalfont,labelsep=colon,strut=off} 
\floatstyle{ruled}
\newfloat{listing}{tb}{lst}{}
\floatname{listing}{Listing}
\title{Breaking Model Lock-in: Cost-Efficient Zero-Shot LLM Routing via a Universal Latent Space}
\newcommand{\systemname}{{ZeroRouter}\xspace}
\author{
    Cheng Yan\textsuperscript{\rm 1},
    Wuyang Zhang\thanks{Corresponding author}\textsuperscript{\rm 1},
    Zhiyuan Ning\textsuperscript{\rm 1},
    Fan Xu\textsuperscript{\rm 1},
    Ziyang Tao\textsuperscript{\rm 1},
    Lu Zhang\textsuperscript{\rm 2},
    Bing Yin\textsuperscript{\rm 3},
    Yanyong Zhang\footnotemark[1]\textsuperscript{\rm 1,\rm 2}
}
\affiliations{
    \textsuperscript{\rm 1}University of Science and Technology of China\\
    \textsuperscript{\rm 2}Institute of Artificial Intelligence, Hefei Comprehensive National Science Center\\
    \textsuperscript{\rm 3}Research Department, iFLYTEK Co., LTD.

    \{yc\_sa22218099, markxu, ziyangtao\}@mail.ustc.edu.cn, 
    \{wuyangz, luzha, yanyongz\}@ustc.edu.cn
}

\usepackage{bibentry}

\begin{document}

\maketitle

\begin{abstract}

The rapid proliferation of Large Language Models (LLMs) has led to a fragmented and inefficient ecosystem, a state of ``model lock-in'' where seamlessly integrating novel models remains a significant bottleneck. Current routing frameworks require exhaustive, costly retraining, hindering scalability and adaptability. We introduce \systemname, a new paradigm for LLM routing that breaks this lock-in. Our approach is founded on a universal latent space, a model-agnostic representation of query difficulty that fundamentally decouples the characterization of a query from the profiling of a model. This allows for zero-shot onboarding of new models without full-scale retraining. \systemname features a context-aware predictor that maps queries to this universal space and a dual-mode optimizer that balances accuracy, cost, and latency. Our framework consistently outperforms all baselines, delivering higher accuracy at lower cost and latency.
\end{abstract}

\begin{links}
    \link{Code}{https://github.com/Codeffun3/ZeroRouter}
\end{links}

\section{Introduction}

The landscape of Large Language Models (LLMs) is no longer monolithic. 
While massive models deliver peak performance, their computational cost makes them inefficient for routine tasks that smaller models could handle. 
Left to users, model selection is often manual and haphazard, leading to expensive failures: over-reliance on costly, oversized models for simple queries, and the underutilization of more efficient alternatives. 
Here, the core challenge is to automate this selection process, making routing systems that can dynamically navigate the trade-offs between cost, latency, and quality.

Current LLM routing frameworks adopt two paradigms with inherent scalability limitations. 
Direct routing systems, such as HybridLLM~\cite{ding2024hybrid} and RouteLLM~\cite{ong2024routellm}, use static model-task mappings or cascading workflows, which require exhaustive query evaluations across predefined models. 
These methods cannot integrate new models without architectural overhauls, as their logic is hardcoded for fixed model sets, and cascading workflows incur quadratic evaluation costs as the model pool grows.
Inference-free predictive frameworks, such as MixLLM~\cite{wang2025mixllm}, uses a contextual-bandit framework to balance quality, cost, and latency and continually adapts to user feedback. When new models are introduced, these predictors fail to capture task-query-LLM interactions critical for generalization.
\begin{figure}[t]
\centering
\includegraphics[width=0.99\linewidth]{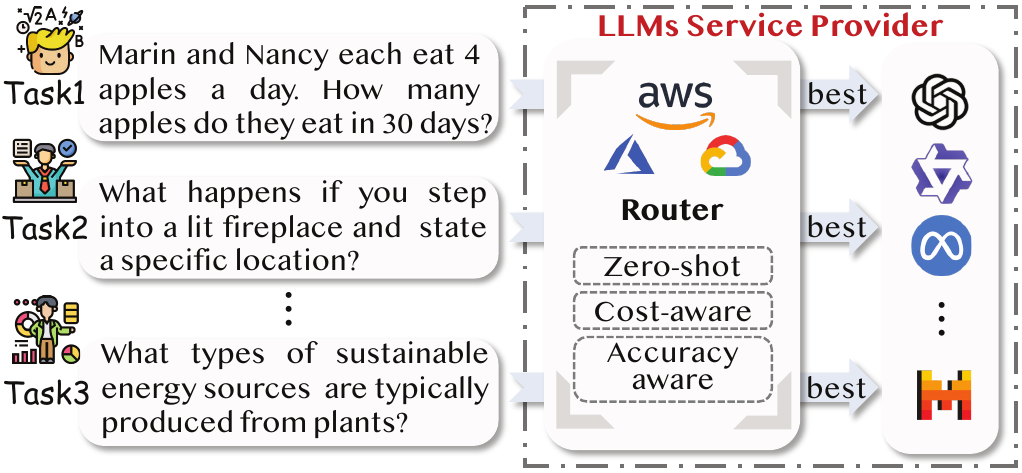}
\caption{Example of Large Language Routing Mechanism. Queries are intelligently routed to the most suitable LLM from service providers for optimal performance and cost.}
\label{fig:intro}
\end{figure}
The primary limitation of most existing LLM routing systems is their static design. 
This creates a state of "model lock-in," where the router is tied to a fixed set of models. Incorporating new LLMs usually requires a complete and costly overhaul, including extensive retraining with new preference data. 
They may generalize poorly to out-of-distribution queries and new tasks because they fail to grasp the intrinsic properties of a query, relying instead on superficial heuristics. 
Furthermore, their performance is often compromised by unreliable signals like miscalibrated model confidence, and they operate without awareness of real-time dynamics of model inference.

This paper introduces \systemname, a framework that rebuilds LLM routing from the ground up to break free from these  limitations. 
We propose a new paradigm centered on a universal, cross-task latent space, a multi-dimensional map that represents the intrinsic characteristics of any query in a model-agnostic way. 
This core innovation fundamentally decouples the task of characterizing a query from the task of profiling a model.
New models can therefore be integrated seamlessly by efficiently charting their performance capabilities onto this universal map using only a small set of anchor queries, eliminating the need for costly, full-scale retraining.
A context-aware predictor then maps raw text directly to a coordinate in this latent space, allowing a lightweight yet highly intelligent routing decision. By framing routing as a multi-objective optimization problem, \systemname provides a scalable and adaptive solution that assigns each query to the most cost-effective mode.
Our framework consistently outperforms all baselines by simultaneously delivering higher accuracy at lower cost and latency. This lead is most significant on unseen datasets, where it achieves a higher maximum accuracy while also dramatically outperforming all rivals in minimizing cost and latency.

\section{Related Work}

\subsection{LLM Routing} 
Existing approaches for LLM routing~\cite{zhang2025capability,wang2025mixllm,aggarwal2024automix,mohammadshahi2024routoo} face critical trade-offs between adaptability and efficiency. Router-trained systems such as HybridLLM~\cite{ding2024hybrid} and RouteLLM~\cite{ong2024routellm} map queries to models via supervised learning, yet their transductive design locks them to predefined model sets where new LLMs demand full router retraining. 
The current frontier is inference-free predictive frameworks(e.g. FORC~\cite{vsakota2024fly}), which decouple performance prediction and optimization. MixLLM~\cite{wang2025mixllm}, uses a contextual-bandit framework to dynamically balance quality, cost, and latency and continually adapts to user feedback. While these methods represent significant progress, they face bottlenecks such as mandatory profiling for new models or reliance on heuristics for unseen queries.

\subsection{Efficient Benchmarking of LLMs}
Previous work reduces the cost of LLM evaluation with strategic sub-sampling. MixEval~\cite{ni2024mixeval} selects subsets to predict aggregate benchmark scores, while TinyBenchmark~\cite{polo2024tinybenchmarks} and MetaBench~\cite{kipnis2024texttt} leverage clustering or Item Response Theory (IRT)~\cite{rodriguez2021evaluation} to identify diagnostic instances. Though these methods enable lightweight profiling, they prioritize dataset-level metric approximation and cannot generalize to unseen prompts. Recent efforts (e.g. How2Bench~\cite{cao2025should}) provides comprehensive guidelines for benchmark development, and the creation of specialized test suites for nuanced abilities. 
However, this creates a scalability paradox: as benchmarks become more numerous and rigorous, the cost to vet new models escalates, reinforcing a paradigm of exhaustive, slow evaluation. Our framework breaks this cycle by establishing a specific-model-agnostic difficulty scale, enabling lightweight profiling of new models without costly, ever-expanding benchmark evaluations.

\section{Methodology}
\begin{figure*}[t]
\centering
\includegraphics[width=1.0\linewidth]{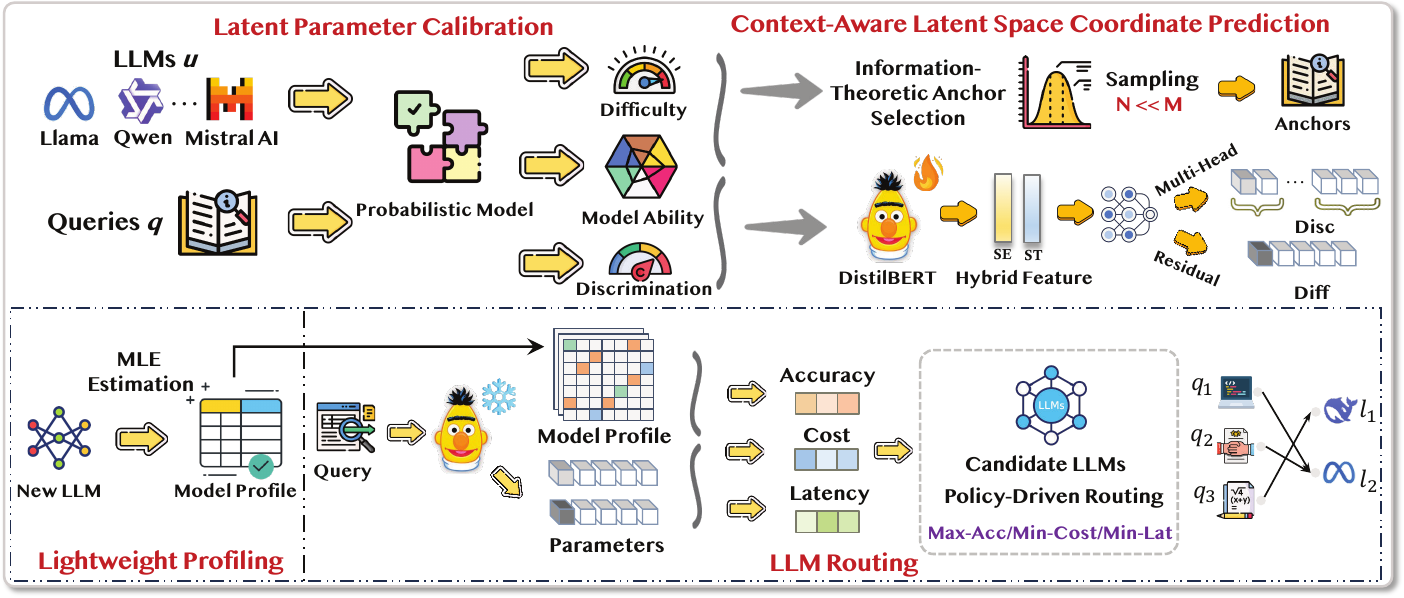}
\caption{
    An overview of the ZeroRouter framework, detailing its three core modules: 
    (1) \textbf{Latent Parameter Calibration}, which establishes a universal latent space using information-theoretic anchor selection; 
    (2) \textbf{Lightweight Profiling}, which efficiently maps new LLMs onto this space; and 
    (3) \textbf{Policy-Driven Routing}, which predicts a query's latent coordinates and assigns it to the optimal model based on user-defined constraints for accuracy, cost, and latency.
}
\label{fig:overview}
\end{figure*}

Our framework pioneers a zero-shot routing paradigm by establishing a context-aware link between a query's intrinsic properties and its performance. 
Unlike prior works that require exhaustive evaluations or full retraining, \systemname achieves scalable model integration through three key innovations: \ding{182} a universal, cross-task latent space, calibrated by an information-theoretic selection strategy, to decouple prompt characteristics from model capabilities; \ding{183} a context-aware latent space predictor that maps a query's textual properties directly to its latent parameters; and \ding{184} a unified policy-driven routing engine that translates high-level user priorities on accuracy, cost, and latency into optimal query assignments. Figure \ref{fig:overview} illustrates the overall framework.

\subsection{Universal Latent Space for Lightweight Profiling}
This module provides lightweight accuracy, cost, and latency profiles for new LLMs by decoupling a prompt's intrinsic latent parameters (e.g., discrimination, difficulty), which are model-agnostic, from a model's specific ability profile. This separation allows highly efficient profiling of new models, requiring performance data only on a small set of anchor prompts to accurately estimate their capabilities.

\subsubsection{Model Inference Accuracy Estimation}\leavevmode

\noindent\textit{\textbf{Cross-Task Discrimination and Difficulty Calibration.}} 
To create a universal discrimination and difficulty scale, we first model the interaction between a model \(u\) and a prompt \(i\) using a Probabilistic Model, such as multidim 2-Parameter Logistic (2PL) IRT model~\cite{rodriguez2021evaluation}. The probability of a correct response \(X_{ui}=1\) is given by:
\begin{equation}
    \label{eq:mirt}
    P(X_{ui}=1 | \boldsymbol{\theta_u}, \boldsymbol{\alpha_i}, \boldsymbol{b_i}) = \frac{1}{1 + e^{-\boldsymbol{\alpha_i}^\top(\boldsymbol{\theta_u} - \boldsymbol{b_i})}},
\end{equation}
where \(\boldsymbol{\theta_u}\) is the model's ability vector, while \(\boldsymbol{\alpha_i}\) and \(\boldsymbol{b_i}\) are the prompt's discrimination and difficulty vectors, respectively, all residing in a \(D\)-dimensional latent space (\(\in \mathbb{R}^D\)). We define a hierarchical Bayesian model over these latent variables (with priors \(\boldsymbol{\theta_u} \sim \mathcal{N}(\boldsymbol{\mu_\theta}, \boldsymbol{u_\theta}^{-1})\), etc.) and perform efficient posterior inference by maximizing the Evidence Lower Bound (ELBO) via Stochastic Variational Inference (SVI). The key innovation is to build a universal scale via cross-task calibration, allowing the derivation of a minimal set of anchors for the efficient evaluation of new models.

\noindent\textit{\textbf{Information-Theoretic Anchor Selection.}}
To select a minimal yet highly informative anchor set $\mathcal{A} \subset \mathcal{P}$ with size $N \ll |\mathcal{P}|$ that effectively estimates model capabilities, we formalize the task as an optimal experimental design problem, with the aim of minimizing the uncertainty of the model's estimated ability parameter $\boldsymbol{\theta}_{u}$. To achieve this, we employ the principle of D-optimality, which seeks to maximize the determinant of the Fisher Information Matrix:
\begin{equation}
\mathcal{I}_{\mathcal{A}}(\boldsymbol{\theta}_u) = \sum_{i \in \mathcal{A}} P(X_{ui}=1)(1-P(X_{ui}=1)) \boldsymbol{\alpha}_i \boldsymbol{\alpha}_i^\top
\label{eq:fisher_info_full}
\end{equation}
Since the probability term in Eq. (\ref{eq:fisher_info_full}) depends on the unknown model ability $\boldsymbol{\theta}_u$, we simplify the objective to rely solely on the prompt discrimination vectors $\boldsymbol{\alpha}_i$, which capture the model-agnostic informativeness of each prompt. The objective is thus to maximize the log-determinant of the cumulative discrimination matrix:
\begin{equation}
\mathcal{A}^* = \underset{\mathcal{A} \subset \mathcal{P}, |\mathcal{A}|=N}{\arg\max} \log \det\left(\sum_{i \in \mathcal{A}} \boldsymbol{\alpha}_i \boldsymbol{\alpha}_i^\top\right)
\label{eq:d_optimal_objective}
\end{equation}
To avoid directly solving the NP-hard optimization problem, we construct the set $\mathcal{A}$ using a greedy forward selection algorithm. Starting with an empty set $\mathcal{A}_0 = \emptyset$ and a regularized matrix $\mathcal{I}_0 = \epsilon\mathbf{I}$ for numerical stability, the algorithm iteratively appends the prompt $i_k^*$ that offers the maximal gain:
\begin{equation}
i_k^* = \underset{i \in \mathcal{P} \setminus \mathcal{A}_{k-1}}{\arg\max} \log \det(\mathcal{I}_{k-1} + \boldsymbol{\alpha}_i \boldsymbol{\alpha}_i^\top)
\label{eq:greedy_selection}
\end{equation}
This process is repeated until $|\mathcal{A}_k| = N$. The resulting set of anchors is compact and highly informative, providing a solid foundation for cost-effective profiling.

\noindent\textit{\textbf{Lightweight Profiling for New Models.}} 
With the information-theoretic anchor set $\mathcal{A}$ established, we can efficiently estimate $\boldsymbol{\theta}_{u_{\text{new}}}$ by minimizing the binary cross-entropy (BCE) loss:
\begin{equation}
\hat{\boldsymbol{\theta}}_{u_{\text{new}}} = \underset{\boldsymbol{\theta}_{u_{\text{new}}}}{\arg\min} \sum_{k \in \mathcal{A}} \BCE(y_k, P_k),
\label{eq:mle_bce}
\end{equation}
where $y_k \in [0, 1]$ is the ground-truth score for anchor $k \in \mathcal{A}$, and $P_k$ is the correctness probability predicted by Eq.~(\ref{eq:mirt}) using the new model's ability $\boldsymbol{\theta}_{u_{\text{new}}}$. The term $\BCE(y, p)$ defined as $-\left[y \ln(p) + (1 - y) \ln(1 - p)\right]$.

\subsubsection{Model Inference Cost Estimation}\leavevmode

\noindent Accurate cost estimation is a prerequisite for effective routing. For API-based LLM services, the total inference cost for routing a query \(q\) to a model \(u\) is a function of its input and output token lengths:
\begin{equation}
    \mathcal{C}_{uq} = \lambda_u^{\mathrm{in}} \ell_{uq}^{\mathrm{in}} + \lambda_u^{\mathrm{out}} \ell_{uq}^{\mathrm{out}},
    \label{eq:single_cost}
\end{equation}
where \( \lambda_u^{\mathrm{in}} \) and \( \lambda_u^{\mathrm{out}} \) are the per-token API prices for model \(u\), while \( \ell_{uq}^{\mathrm{in}} \) and \( \ell_{uq}^{\mathrm{out}} \) are the input and output token lengths. 

\noindent\textit{\textbf{Input Token Estimation.}} 
Estimating the input token length, $\ell^{\mathrm{in}}$, is deterministic. Given query $q$ and model $u$, we employ the model-specific tokenizer, $\mathcal{T}_u(\cdot)$, to obtain a precise count:
\begin{equation}
    \ell_{uq}^{\mathrm{in}} = |\mathcal{T}_u(q)|.
\end{equation}

\noindent\textit{\textbf{Output Token Estimation.}}
Unlike the deterministic input length, estimating the output token length $\ell_{uq}^{\text{out}}$ is non-trivial, as it depends on both query complexity and the specific model's verbosity. We introduce a novel estimation method that directly leverages our universal latent space, bypassing the need for complex, per-model regression heads for this task. 
First, we define the task-aware difficulty $s_q$, derived from the predicted latent IRT parameter vectors:
\begin{equation}
s_q = \boldsymbol{\alpha}_q^\top \boldsymbol{b}_q
\label{eq:complexity_score}
\end{equation}
During a one-time calibration phase using our small anchor set $\mathcal{A}$, we discretize the continuous range of complexity scores observed across the anchor samples into $K$ bins, $\mathcal{B} = \{B_1, \dots, B_K\}$. 
For each candidate model $u \in \mathcal{M}$ and  complexity bin $B_k$, we pre-compute and store a mean output token length, $\bar{\ell}_{uk}^{\text{out}}$, by averaging over the ground-truth lengths of anchor samples whose complexity scores fall into that bin:
\begin{equation}
\bar{\ell}_{uk}^{\text{out}} = \mathbb{E}_{q \in \mathcal{A} \,|\, s_q \in B_k} \left[ |\mathcal{T}_u(\text{response}(q, u))| \right]
\label{eq:bin_averaging}
\end{equation}
This process generates a lightweight lookup table mapping a $(\text{model}, \text{complexity\_bin})$ tuple to an expected output length, efficiently profiling each model's verbosity across different complexity tiers.

For an unseen query $q'$, the estimation process is remarkably efficient. We first predict its latent vectors $(\hat{\boldsymbol{\alpha}}_{q'}, \hat{\boldsymbol{b}}_{q'})$ using our context-aware predictor. 
We then compute its estimated complexity score $\hat{s}_{q'} = \hat{\boldsymbol{\alpha}}_{q'}^\top \hat{\boldsymbol{b}}_{q'}$, identify the corresponding bin $B_k$ such that $\hat{s}_{q'} \in B_k$, and simply retrieve the pre-calibrated average length for the target model $u$:
\begin{equation}
\hat{\ell}_{uq'}^{\text{out}} = \bar{\ell}_{uk}^{\text{out}}
\label{eq:length_lookup}
\end{equation}
This decoupling of latent parameters from model execution supports zero-shot generalization, reducing output length estimation for any new query to a inference-free table lookup.
\subsubsection{Model Inference Latency Estimation}\leavevmode

\noindent We model latency for each model $u$ via two pre-calibrated parameters: its Time to First Token ($\tau_{u}^{\text{TTFT}}$), capturing initial processing overhead, and its Time per Output Token ($\tau_{u}^{\text{TPOT}}$), reflecting its generation speed. These parameters are efficiently estimated for each model using the anchor set $\mathcal{A}$. The final latency for a query $q$ is predicted by:
\begin{equation}
    \hat{\tau}_{uq} = \tau_{u}^{\text{TTFT}} + \hat{\ell}_{uq}^{\text{out}} \cdot \tau_{u}^{\text{TPOT}}.
    \label{eq:latency_model}
\end{equation}
This approach provides an efficient latency estimation, grounded in the fundamental mechanics of LLM inference.

\subsection{Context-Aware Latent Space Coordinate Prediction}
To support effective routing generalization to out-of-distribution (OOD) queries, we introduce a lightweight unified predictor that bridges a query's intrinsic textual properties with its multi-dimensional IRT parameter vectors: the discrimination vector $\boldsymbol{\alpha} \in \mathbb{R}^D$ and the difficulty vector $\boldsymbol{b} \in \mathbb{R}^D$.

\subsubsection{Hybrid Feature Representation}\leavevmode

\noindent Our approach posits that a query's latent parameters arise from both its semantic depth and structural complexity. We encode a given query $q$ into following two feature vectors:

\noindent\textit{\textbf{Semantic Embedding ($\boldsymbol{e}_{\mathrm{se}}$).}} We capture contextual meaning using a fine-tuning DistilBERT model to extract the final hidden state of the [CLS] token.
    \begin{equation}
        \boldsymbol{e}_{\mathrm{se}} = \mathrm{DistilBERT}(q)_{[\mathrm{CLS}]} \in \mathbb{R}^{d_{\mathrm{sem}}}
    \end{equation}
\noindent\textit{\textbf{Structural Features ($\boldsymbol{e}_{\mathrm{st}}$).}}We quantify surface-level complexity using a feature extraction function $\Phi(\cdot)$, which computes $k=11$ linguistic metrics (e.g., readability scores, parse tree depth). The selection of these hybrid features is guided by a correlation analysis with the target IRT parameters, ensuring they provide relevant predictive signals.
    \begin{equation}
        \boldsymbol{e}_{\mathrm{st}} = \Phi(q) \in \mathbb{R}^{k}
    \end{equation}

\subsubsection{Unified Prediction Model}\leavevmode

\noindent To co-predict both $\boldsymbol{\alpha}$ and $\boldsymbol{b}$ from these features, we designed a multi-task learning model, This model consists of a shared backbone that learns a high-level representation of the query, followed by two independent prediction heads.

\noindent\textit{\textbf{Shared Backbone.}} The backbone first projects each feature vector into a higher-dimensional space using dedicated linear layers, then fuses them through a deep non-linear trunk network $f_{\mathrm{fuse}}$ for robust feature learning. This produces the shared latent representation:
\begin{equation}
    \boldsymbol{h}_{\mathrm{shared}} = f_{\mathrm{fuse}}\left(\left[\mathbf{W}_{\mathrm{se}}\boldsymbol{e}_{\mathrm{se}} + \boldsymbol{e}_{\mathrm{se}} \ ; \ \mathbf{W}_{\mathrm{st}}\boldsymbol{e}_{\mathrm{st}} + \boldsymbol{b}_{\mathrm{st}}\right]\right),
\end{equation}
where $(\mathbf{W}_k, \boldsymbol{b}_k)$ are trainable parameters for $k \in \{\mathrm{se}, \mathrm{st}\}$.

\noindent\textit{\textbf{Difficulty Head (Residual Prediction).}} Predicting the absolute position of the difficulty vector $\boldsymbol{b}$ across a high-dimensional space is challenging. However, we observe that a query's difficulty vector often represents a small deviation from a data-driven mean $\boldsymbol{\bar{b}}$ computed across the training sets. We therefore reformulate the task to predict only this \textit{residual} vector, which significantly simplifies the learning problem. The final predicted difficulty vector $\hat{\boldsymbol{b}}_q$ is constructed as:
\begin{equation}
    \hat{\boldsymbol{b}}_q = \boldsymbol{\bar{b}} + \Delta\boldsymbol{b}_q = \boldsymbol{\bar{b}} + f_{\mathrm{diff}}(\boldsymbol{h}_{\mathrm{shared}}),
\end{equation}
where $f_{\mathrm{diff}}$ is the difficulty head, implemented as a multi-layer perceptron (MLP), that takes the shared latent representation $\boldsymbol{h}_{\mathrm{shared}}$ as input to predict the residual vector $\Delta\boldsymbol{b}_q$.

\noindent\textit{\textbf{Discrimination Head (Multi-Head Experts).}} The dimensions of the discrimination vector $\boldsymbol{\alpha}$ are often correspond to related task types or specific skills. A monolithic prediction head struggles to capture these nuanced inter-dimensional relationships. We therefore partition the $D$ dimensions into $C$ distinct clusters based on inter-dimensional correlation analysis.  For each cluster, a dedicated expert head specializes in predicting only the subset of dimensions within its assigned group. The final discrimination vector $\hat{\boldsymbol{\alpha}}_q$ is then constructed by concatenating the outputs from each expert:
\begin{equation}
    \hat{\boldsymbol{\alpha}}_q = \bigoplus_{c=1}^{C} \hat{\boldsymbol{\alpha}}_{q,c} = \bigoplus_{c=1}^{C} f_{\mathrm{disc},c}(\boldsymbol{h}_{\mathrm{shared}}),
\end{equation}
where $\bigoplus$ denotes the concatenation and re-ordering of the vector outputs from each expert head, and each expert head $f_{\mathrm{disc},c}$ is a dedicated multi-layer perceptron (MLP) mapping the shared representation $\boldsymbol{h}_{\mathrm{shared}}$ to its corresponding discrimination sub-vector $\hat{\boldsymbol{\alpha}}_{q,c}$.

\noindent By estimating these latent vectors, we can project how one model will perform on an unseen query, which is the key to our zero-shot routing capability.

\subsection{LLM Routing Strategies}
Effective LLM routing requires navigating a complex, multi-dimensional trade-off between accuracy, cost, and latency. To address this, \systemname introduces a unified optimization framework that allows developers to flexibly express their priorities within a coherent system. 
\paragraph{Multi-Objective Optimization.}
We formulate the routing task as a Integer Linear Programming (ILP) problem that maximizes a weighted utility function, unifying the competing goals of accuracy, cost, and latency. For a set of queries $\mathcal{Q}$ and a pool of models $\mathcal{M}$, we define the predicted accuracy $p_{uq}$, estimated cost $\mathcal{C}_{uq}$, and estimated latency $\tau_{uq}$ for each query-model pair $(u, q)$. Our goal is to determine the optimal assignment, represented by a binary variable $x_{uq} \in \{0, 1\}$, guided by three user-specified weights ($w_p, w_c, w_t$) that sum to 1. The objective is formulated as:
\begin{equation}
    \operatorname{maximize}~\sum_{u \in \mathcal{M}} \sum_{q \in \mathcal{Q}} (w_p p_{uq} - w_c \mathcal{C}_{uq} - w_t \tau_{uq}) \cdot x_{uq},
    \label{eq:objective}
\end{equation}
subject to the fundamental constraint that each query is assigned to exactly one model, and a set of optional global constraints on total cost, latency and average accuracy:
\begin{equation}
\label{eq:ilp_constraints}
\begin{aligned}
    \text{subject to} ~ & \sum_{u \in \mathcal{M}} x_{uq} = 1, && \forall q \in \mathcal{Q} \\
    & \sum_{u \in \mathcal{M}} \sum_{q \in \mathcal{Q}} r_{uq} \cdot x_{uq} \le R_{\text{max}}, && \text{for } r \in \{\mathcal{C}, \tau\} \\
    & \sum_{u \in \mathcal{M}} \sum_{q \in \mathcal{Q}} p_{uq} \cdot x_{uq} \ge |\mathcal{Q}| \cdot p_{\text{min}}. &&
\end{aligned}
\end{equation}
\paragraph{Policy-Driven Routing.}
This unified formulation is highly expressive. By adjusting the weights $(w_p, w_c, w_t)$, developers can define a wide spectrum of routing behaviors without any change to the underlying system architecture:
\ding{182} \textit{\textbf{Accuracy-First ($w_p \to 1$)}}: For critical applications, maximizing accuracy is paramount.
\ding{183} \textit{\textbf{Cost-First ($w_c \to 1$)}}: For budget-constrained services, the router prioritizes the cheapest models.
\ding{184} \textit{\textbf{Latency-First ($w_t \to 1$)}}: For interactive applications, minimizing response time is crucial.
\ding{185} \textit{\textbf{Balanced Strategy}}: Our evaluation adopts a mix of weights (e.g., $w_p=0.5, w_c=0.3, w_t=0.2$) that find a balance case.

\section{Experiments}

\subsection{Setup and Implementation Details}
\label{Experimental Setup}

\noindent\textbf{Datasets.}
We evaluate \systemname with nine datasets, divided into six in-distribution (ID) sets to measure performance on familiar tasks, and three out-of-distribution (OOD) sets not seen in training to test its zero-shot generalization.

\noindent\textit{\textbf{Six ID benchmarks:}}
\textbf{(1) IFEval} \cite{zhou2023instruction} for evaluating instruction following fidelity, \textbf{(2) BBH} (Big-Bench Hard) \cite{suzgun2022challenging} for tasks requiring multi-step reasoning, \textbf{(3) MATH} \cite{hendrycks2021measuring} for advanced mathematical problem solving, \textbf{(4) GPQA} (Graduate-Level Google-Proof Q\&A) \cite{rein2024gpqa} for difficult questions, \textbf{(5) MuSR} (Multi-Step Reasoning) \cite{yao2025mmreason} for its focus on sequential reasoning, and \textbf{(6) MMLU-PRO} \cite{wang2024mmlu}, a programmatically-verified version of MMLU. 

\noindent\textit{\textbf{Three OOD benchmarks:}}
\textbf{(1) ARC-C} (AI2 Reasoning Challenge) \cite{clark2018think}, a collection of science questions, \textbf{(2) TruthfulQA} \cite{lin2021truthfulqa}, which measures a model's propensity to avoid generating falsehoods, and \textbf{(3) HumanEval} \cite{li2024humaneval} for evaluating code generation capabilities. 

\noindent\textbf{LLM Candidates.}
We construct evaluations on a curated pool of 60 Large Language Models (LLMs) strategically divided into two sets.
A core set of 10 models, with parameter counts spanning from a compact 1B to a massive 235B, was used to test routing performance across a diverse and challenging landscape:
Qwen3-235B-A22B, DeepSeek-R1-Distill-Llama-70B, Mixtral-8x7B-Instruct-v0.1 \cite{jiang2024mixtral}, Qwen3-32B, gemma-3-27b-it, phi-4, Llama-3.1-8B-Instruct, DeepSeek-R1-0528-Qwen3-8B, gemma-3n-E4B-it, and gemma-3-1b-it.
A second cohort of 50 distinct LLMs being released after our router's training data cutoff was used to directly validate our central claim of breaking model lock-in. 

\noindent\textbf{Baselines.}
We compare our framework against four recent representative works: (1) \textbf{CIT-LLM-Routing} \cite{zhang2025capability}, which introduces a novel paradigm of capability instruction tuning based on model aptitude tests to predict performance. (2) \textbf{RouteLLM} \cite{ong2024routellm}, which adopts a binary routing strategy that uses human preference data to assign queries to either a "strong" or "weak" model. (3) \textbf{GraphRouter} \cite{feng2024graphrouter}, a method that constructs a heterogeneous graph to explicitly model interactions among tasks, queries, and LLMs, reframing routing as an edge prediction problem. (4) \textbf{FORC} \cite{vsakota2024fly}, which employs a two-stage routing method that first uses a pre-trained meta-model to predict query accuracy for each LLM, and then selects the most suitable model based on these predictions.

\noindent\textbf{Evaluation Metrics.}
To comprehensively evaluate the trade-offs of each router, we define a unified total reward over the test query set $\mathcal{Q}_{\text{test}}$ as:
\begin{equation}
    \text{Reward} = \sum_{q \in \mathcal{Q}_{\text{test}}} (w_p p_{u^*q} - w_c \mathcal{C}_{u^*q} - w_t \tau_{u^*q}),
    \label{eq:reward_metric}
\end{equation}
where $u^*$ is the model selected by the router for query $q$. The components $p_{uq}$, $\mathcal{C}_{uq}$, and $\tau_{uq}$ are the accuracy, cost, and latency previously defined for a query-model pair. Specifically, accuracy is evaluated against ground truth (see the Appendix for more details), cost is calculated as defined in Eq.~\ref{eq:single_cost}, and latency is end-to-end inference time.

Our evaluation considers three policy-driven scenarios with respective weights ($w_p, w_c, w_t$): Accuracy-First (\textbf{Max-Acc}) $(0.8, 0.1, 0.1)$, Cost-First (\textbf{Min-Cost}) $(0.1, 0.8, 0.1)$, and Latency-First (\textbf{Min-Lat}) $(0.1, 0.1, 0.8)$.

\begin{table*}[t]
\small
\centering
\setlength\tabcolsep{2pt}

\begin{tabular}{ p{2.6em} p{6.2em}<{\centering} p{4.8em}<{\centering} p{4.7em}<{\centering}  p{4.7em}<{\centering}  p{4.7em}<{\centering}  p{4.7em}<{\centering}  p{4.7em}<{\centering}  p{4.7em}<{\centering}  p{4.7em}<{\centering}}
\toprule

\multirow{2}{*}{\textbf{Method}} & & \multirow{2}{*}{\textbf{\#Params}} & \multicolumn{3}{c}{\textbf{In-Domain}} & \multicolumn{3}{c}{\textbf{Out-of-Domain}} & \multirow{2}{*}{\textbf{Mean}} \\
& & & Max-Acc & Min-Cost & Min-Lat &Max-Acc & Min-Cost & Min-Lat & \\
\midrule
& & & & & \multicolumn{5}{l}{{\em \quad \quad \, Smaller-scale LLMs ($<$20B)}} \\
\noalign{\vskip 0.7ex}
\cline{7-9}
\noalign{\vskip 0.7ex}
\multicolumn{2}{l}{Gemma-3} & 1B & 0.15 & -0.38 & -0.30 & 0.44 & -0.28 & -0.24 & -0.10 \\ 
\multicolumn{2}{l}{Gemma-3n-E4B} & 7.8B & 0.25 & -0.36 & -0.41 & 0.47 & -0.25 & -0.41 & -0.12 \\ 
\multicolumn{2}{l}{DeepSeek-R1-0528-Qwen3} & 8B & 0.33 & -0.58 & -0.71 & 0.58 & -0.56 & -0.71 & -0.28 \\ 
\multicolumn{2}{l}{Llama-3.1$_{\text{\ Instruct}}$} & 8B & 0.29 & -0.54 & -0.65 & 0.59 & -0.45 & -0.66 & -0.24 \\
\multicolumn{2}{l}{Phi-4} & 14.7B & 0.41 & -0.82 & -0.71 & 0.55 & -0.83 & -0.78 & -0.36 \\

\noalign{\vskip 0.7ex}
& & & \multicolumn{7}{l}{\quad \quad \quad \quad \quad { \textbf{LLM Selection} on \textit{Smaller-scale} LLMs (5 models)}} \\

\noalign{\vskip 0.7ex}
\cline{7-9}
\noalign{\vskip 0.7ex}
\multicolumn{2}{l}{{Random Selection}} & $-$ & 0.29\rlap{\textsuperscript{\textdagger}} & -0.54\rlap{\textsuperscript{\textdagger}} & -0.55\rlap{\textsuperscript{\textdagger}} & 0.52\rlap{\textsuperscript{\textdagger}} & -0.51\rlap{\textsuperscript{\textdagger}} & -0.55\rlap{\textsuperscript{\textdagger}} & -0.22 \\
\multicolumn{2}{l}{RouteLLM} & $-$ & 0.30\rlap{\textsuperscript{\textdagger}} & -0.47\rlap{\textsuperscript{\textdagger}} & -0.53\rlap{\textsuperscript{\textdagger}} & 0.51\rlap{\textsuperscript{\textdagger}} & -0.54\rlap{\textsuperscript{\textdagger}} & -0.48\rlap{\textsuperscript{\textdagger}} & -0.20 \\
\multicolumn{2}{l}{FORC} & $-$ & 0.35\rlap{\textsuperscript{\textdagger}} & -0.36\rlap{\textsuperscript{\textdagger}} & -0.54\rlap{\textsuperscript{\textdagger}} & 0.50\rlap{\textsuperscript{\textdagger}} & -0.57\rlap{\textsuperscript{\textdagger}} & -0.54\rlap{\textsuperscript{\textdagger}} & -0.19 \\
\multicolumn{2}{l}{GraphRouter} & $-$ & 0.38\rlap{\textsuperscript{\textdagger}} & -0.40\rlap{\textsuperscript{\textdagger}} & -0.56\rlap{\textsuperscript{\textdagger}} & 0.53\rlap{\textsuperscript{\textdagger}} & -0.28\rlap{\textsuperscript{\textdagger}} & -0.50\rlap{\textsuperscript{\textdagger}} & -0.14 \\
\multicolumn{2}{l}{Model-SAT} & $-$ & 0.43 & -0.61\rlap{\textsuperscript{\textdagger}} & -0.64\rlap{\textsuperscript{\textdagger}} & 0.57 & -0.59\rlap{\textsuperscript{\textdagger}} & -0.64\rlap{\textsuperscript{\textdagger}} & -0.25 \\
\multicolumn{2}{l}{\textsc{\systemname} \textbf{(Ours)}} & $-$ & \textbf{0.45} & \textbf{-0.32} & \textbf{-0.27} & \textbf{0.59} & \textbf{-0.24} & \textbf{-0.23} & \textbf{0.00}\\
\bottomrule
\noalign{\vskip 0.7ex}
\noalign{\vskip 0.7ex}
& & & & & \multicolumn{5}{l}{{\em \quad \quad Larger-Scale LLMs (25B\textasciitilde250B)}} \\
\noalign{\vskip 0.7ex}
\cline{7-9}
\noalign{\vskip 0.7ex}
\multicolumn{2}{l}{Gemma-3} & 27B & 0.36 & -0.31 & -0.37 & 0.43 & -0.54 & -0.40 & -0.14 \\
\multicolumn{2}{l}{Qwen3} & 32B & 0.49 & -0.33 & -0.30 & 0.63 & -0.32 & -0.29 & -0.02 \\
\multicolumn{2}{l}{Mixtral-8x7B$_{\text{\ Instruct}}$} & 56B & 0.32 & -0.30 & -0.36 & 0.67 & -0.42 & -0.38 & -0.08 \\ 
\multicolumn{2}{l}{DeepSeek-R1-Distill-Llama} & 70B & 0.38 & -0.55 & -0.35 & 0.58 & -0.68 & -0.52 & -0.19 \\ 
\multicolumn{2}{l}{Qwen3-235B-A22B$_{\text{\ Instruct-v0.1}}$} & 235B & 0.47 & -0.81 & -0.83 & 0.64 & -0.75 & -0.59 & \underline{-0.31}\\

\noalign{\vskip 0.7ex}

& & & \multicolumn{7}{l}{\quad \quad \quad \quad \quad{\textbf{LLM Selection} on \textit{Large-scale} LLMs (5 models)}} \\
\noalign{\vskip 0.7ex}

\cline{7-9}
\noalign{\vskip 0.7ex}
\multicolumn{2}{l}{{Random Selection}} &$-$ & 0.39\rlap{\textsuperscript{\textdagger}} & -0.44\rlap{\textsuperscript{\textdagger}} & -0.45\rlap{\textsuperscript{\textdagger}} & 0.59\rlap{\textsuperscript{\textdagger}} & -0.46\rlap{\textsuperscript{\textdagger}} & -0.44\rlap{\textsuperscript{\textdagger}} & -0.14 \\
\multicolumn{2}{l}{RouteLLM} & $-$ & 0.46\rlap{\textsuperscript{\textdagger}} & -0.45\rlap{\textsuperscript{\textdagger}} & -0.58\rlap{\textsuperscript{\textdagger}} & 0.56\rlap{\textsuperscript{\textdagger}} & -0.47\rlap{\textsuperscript{\textdagger}} & -0.42\rlap{\textsuperscript{\textdagger}} & -0.15 \\
\multicolumn{2}{l}{FORC} & $-$ & 0.49\rlap{\textsuperscript{\textdagger}} & -0.38\rlap{\textsuperscript{\textdagger}} & -0.55\rlap{\textsuperscript{\textdagger}} & 0.58\rlap{\textsuperscript{\textdagger}} & -0.51\rlap{\textsuperscript{\textdagger}} & -0.43\rlap{\textsuperscript{\textdagger}} & -0.13 \\
\multicolumn{2}{l}{GraphRouter} & $-$ & 0.50 & -0.27\rlap{\textsuperscript{\textdagger}} & -0.62\rlap{\textsuperscript{\textdagger}} & 0.61\rlap{\textsuperscript{\textdagger}} & -0.37\rlap{\textsuperscript{\textdagger}} & -0.27\rlap{\textsuperscript{\textdagger}} & -0.07 \\
\multicolumn{2}{l}{Model-SAT} & $-$ & 0.49 & -0.61\rlap{\textsuperscript{\textdagger}} & -0.57\rlap{\textsuperscript{\textdagger}} & 0.62\rlap{\textsuperscript{\textdagger}} & -0.58\rlap{\textsuperscript{\textdagger}} & -0.46\rlap{\textsuperscript{\textdagger}} & -0.19 \\
\multicolumn{2}{l}{\textsc{\systemname} \textbf{(Ours)}} & $-$ & \textbf{0.52} & \textbf{-0.22} & \textbf{-0.28} & \textbf{0.68} & \textbf{-0.17} & \textbf{-0.25} & \textbf{0.05}\\
\bottomrule
\end{tabular}
\caption{
    Comparison of routing performance on In-Distribution (ID) and Out-of-Distribution (OOD) datasets. 
    ZeroRouter consistently outperforms baseline methods across both smaller and larger model scales, and across all routing policies. \textbf{\textsuperscript{\textdagger}} indicates the improvement of \systemname over the baseline is statistically significant ($p < 0.05$) via a Wilcoxon signed-rank test.
}
\label{tab:main_results}
\label{tab:m_res}
\end{table*}

\noindent\textbf{Implementation Details.}
We implement our multidim 2PL IRT model using the \texttt{py-irt} library \cite{natesan2016bayesian} on evaluation data for our ID datasets, which was aggregated from ~200 models on the Open LLM Leaderboard~\cite{open-llm-leaderboard}. We set the latent dimension $D=20$ and train for 6,000 epochs using the Adam optimizer with an initial learning rate of 0.1, which is decayed exponentially by a factor of 0.99 per 100 epochs. The latent parameter predictor, a \texttt{DistilBERT-base-uncased} model (66M), is fine-tuned for 40 epochs with a batch size of 32 and a constant learning rate of 3e-5. The entire training process is conducted on a single NVIDIA A800 GPU (80GB). 

\subsubsection{Performance on In-Distribution Data}

\noindent Our evaluation on in-distribution data (Table~\ref{tab:main_results}) reveals that \systemname fundamentally redefines the cost-performance frontier. 
Across both smaller ($<$20B) and larger (25B-250B) models, our framework consistently outperforms all baselines not just on one objective, but on all simultaneously. 
For instance, with smaller models, \systemname achieves a higher maximum accuracy score 0.45 than the strongest baseline while also securing the best scores for minimum cost -0.32 and latency -0.27. 
This pattern of dominance extends to large-scale models.
This is more than a simple trade-off. 
\systemname consistently achieves higher accuracy at a lower cost and latency, showing the adaptability and effectiveness of our routing solution across diverse scenarios. 

\subsubsection{Zero-Shot Generalization on OOD Datasets}

\noindent The framework's true power is revealed in its zero-shot generalization to OOD datasets, a critical test of its ability to perform on entirely unseen tasks (Table~\ref{tab:main_results}). In this challenging scenarios, \systemname's performance becomes more pronounced. For both smaller and larger models, it not only achieves the highest Max-Acc (e.g., 0.68 vs the strongest baseline's 0.62 for large models) but also dramatically outperforms all rivals on Min-Cost -0.17 and Min-Lat -0.25. 
This consistent success demonstrates that our routing logic avoids the trap of learning superficial, task-specific heuristics. Instead, by modeling the intrinsic properties of a query, \systemname makes reliably informed decisions in new domains, showing the effectiveness and robust generalization of our approach. 

\begin{table}[t]
\small
\label{tab:exp_new_model}
\centering
\setlength\tabcolsep{1.7pt}
\begin{tabular}{ p{2.6em} p{6.2em}<{\centering} p{3.5em}<{\centering} p{3.5em}<{\centering}  p{3.5em}<{\centering} p{3.5em}<{\centering}}
\toprule
\multicolumn{2}{l}{\textbf{Method}} & {\scriptsize \textbf{Max-Acc}} & {\scriptsize \textbf{Min-Cost}} & {\scriptsize \textbf{Min-Lat}} & {\scriptsize \textbf{Mean}} \\
\midrule
\multicolumn{2}{l}{RouteLLM} & 0.28\rlap{\textsuperscript{\textdagger}} & -0.61\rlap{\textsuperscript{\textdagger}} & -0.62\rlap{\textsuperscript{\textdagger}} & -0.32\rlap{\textsuperscript{\textdagger}} \\
\multicolumn{2}{l}{FORC} & 0.28\rlap{\textsuperscript{\textdagger}} & -0.54\rlap{\textsuperscript{\textdagger}} & -0.57\rlap{\textsuperscript{\textdagger}} & -0.28\rlap{\textsuperscript{\textdagger}}  \\
\multicolumn{2}{l}{GraphRouter} & 0.33\rlap{\textsuperscript{\textdagger}} & -0.42\rlap{\textsuperscript{\textdagger}} & -0.44\rlap{\textsuperscript{\textdagger}} & -0.18\rlap{\textsuperscript{\textdagger}} \\
\multicolumn{2}{l}{Model-SAT} & 0.38 & -0.52\rlap{\textsuperscript{\textdagger}} & -0.45\rlap{\textsuperscript{\textdagger}} & -0.20\rlap{\textsuperscript{\textdagger}} \\
\multicolumn{2}{l}{\textsc{\systemname}} &  &  &  &  \\
\multicolumn{2}{l}{\,\,\,\,+ {\scriptsize Random Sampling}} & 0.27\rlap{\textsuperscript{\textdagger}} & -0.56\rlap{\textsuperscript{\textdagger}} & -0.47\rlap{\textsuperscript{\textdagger}} & -0.25\rlap{\textsuperscript{\textdagger}} \\
\multicolumn{2}{l}{\,\,\,\,+ {\scriptsize Diff-based}} & 0.24\rlap{\textsuperscript{\textdagger}} & -0.68\rlap{\textsuperscript{\textdagger}} & -0.52\rlap{\textsuperscript{\textdagger}} & -0.32\rlap{\textsuperscript{\textdagger}} \\
\multicolumn{2}{l}{\,\,\,\,+ {\scriptsize Disc-based}} & 0.28\rlap{\textsuperscript{\textdagger}} & -0.58\rlap{\textsuperscript{\textdagger}} & -0.56\rlap{\textsuperscript{\textdagger}} & -0.29\rlap{\textsuperscript{\textdagger}} \\
\multicolumn{2}{l}{\,\,\,\,+ {\scriptsize Task-aware difficulty}} & 0.32\rlap{\textsuperscript{\textdagger}} & -0.46\rlap{\textsuperscript{\textdagger}} & -0.44\rlap{\textsuperscript{\textdagger}} & -0.19\rlap{\textsuperscript{\textdagger}} \\
\multicolumn{2}{l}{\,\,\,\,+ {\scriptsize D-optimality}} & \textbf{0.39} & \textbf{-0.34} & \textbf{-0.37} & \textbf{-0.11} \\ 
\bottomrule
\end{tabular}

\caption{Ablation study on anchor sampling strategies for efficiently onboarding new models. We compare our proposed D-optimality against alternative sampling strategies and representative baselines. \textbf{\textsuperscript{\textdagger}}~indicates that the improvement of D-optimality over the corresponding method is statistically significant ($p < 0.05$) via a Wilcoxon signed-rank test.}
\label{tab:anchor_ablation}
\end{table}

\subsection{Seamless Integration of New Models}

\subsubsection{Efficient Onboarding with Minimal Data}

To validate our framework's ability to integrate a new model with minimal overhead, we profiled it using a scant budget of just 200 queries (1\% of the training-visible data). The results (Table~\ref{tab:anchor_ablation}) show \systemname decisively outperforms all baselines across every policy. With its core D-optimality sampling, it achieves a superior Max-Acc score (0.39 vs. the baseline best of 0.38 from Model-SAT) and secures the top scores for both Min-Cost (-0.34 from GraphRouter) and Min-Lat (-0.37 from GraphRouter). 
An internal ablation study further underscores the power of our information-theoretic approach, revealing that D-optimality dramatically outperforms simpler strategies like random sampling, boosting Max-Acc from 0.27 to 0.39.

\begin{figure*}[t]
    \centering
    \includegraphics[width=0.85\linewidth]{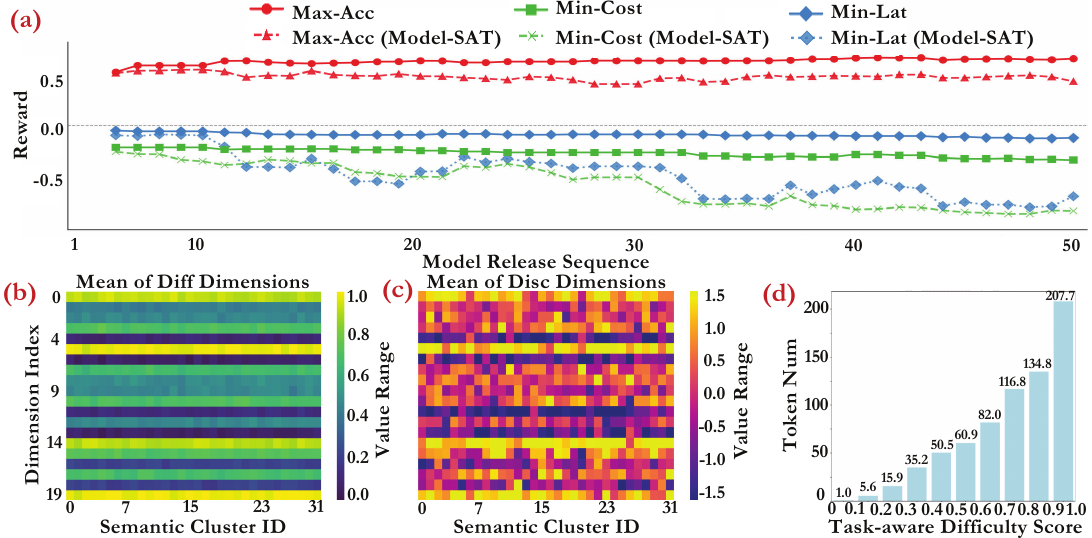}
    \caption{
    Experimental analysis of ZeroRouter's performance and the interpretability of its latent space. 
    \textbf{(a)}~Real-world simulation demonstrating ZeroRouter's ability to seamlessly integrate newly released models and improve performance over time without retraining. 
    \textbf{(b)}~Heatmap of the difficulty vector $\boldsymbol{b}$, showing its task-agnostic nature across different semantic clusters. 
    \textbf{(c)}~Heatmap of the discrimination vector $\boldsymbol{\alpha}$, highlighting its task-specific variability. 
    \textbf{(d)}~Validation of our proposed task-aware difficulty, showing a strong monotonic correlation with the average model output token length.
}
\label{fig:analysis_and_real_world}
\end{figure*}

\subsubsection{Robustness in Real-world Evolving Model Pool}
We further validated our framework in a real-world simulation of rapid model iteration, maintaining a fixed-size pool (N=6) where newly released models sequentially replaced underperforming ones. As depicted in Figure~\ref{fig:analysis_and_real_world}(a), the system intelligently leveraged these more capable models, driving a gradual performance increase under the Max-Accuracy policy while successfully upholding cost and latency constraints when those objectives were prioritized. 
Importantly, our router was trained on data that predates the release of every model in this evolving pool. 

\subsection{Ablation Studies and Analysis}

\subsubsection{Analysis of the Latent Space}

To understand our framework's effectiveness, we analyze the distinct roles of the Difficulty ($\boldsymbol{b}$) and Discrimination ($\boldsymbol{\alpha}$) vectors, shown in Figure~\ref{fig:analysis_and_real_world}. 
This analysis reveals how the latent space separates task-agnostic complexity from task-specific relevance.

\paragraph{The Nature of Difficulty ($\boldsymbol{b}$).}
The heatmap in Figure~\ref{fig:analysis_and_real_world}(b)  offers an interesting insight into the nature of our latent space. Evidenced by uniform horizontal bands, the Difficulty vector $\boldsymbol{b}$, remains remarkably consistent across all semantic task clusters for any given dimension. 
This suggests that $\boldsymbol{b}$ does not represent task-specific difficulty, but rather captures a universal, task-agnostic challenge inherent to each latent dimension. For example, the consistently high value along dimension $b_{19}$  suggests it corresponds to a fundamentally demanding latent capability, such as complex reasoning, which is a prerequisite for success accross a wide variety of tasks. 

\paragraph{The Role of Task-Specific Discrimination ($\boldsymbol{\alpha}$).}
In stark contrast, the Discrimination vector $\boldsymbol{\alpha}$, in Figure~\ref{fig:analysis_and_real_world}(c) is highly variable across different task clusters. This indicates that $\boldsymbol{\alpha}$ captures task-specific properties, quantifying how discriminative a given query is for each latent ability. For instance, dimension $\alpha_{19}$ shows high values primarily for mathematical and logical reasoning tasks, confirming this dimension is indeed associated with that ability. Furthermore, we observe that certain groups of dimensions co-vary across tasks (e.g., dimensions {4, 6, 11, 13} show similar patterns), suggesting they form "ability clusters" representing related skills."

\paragraph{Validation of Task-Aware Difficulty ($\boldsymbol{s_q}$)}
The validity of our task-aware difficulty metric, derived from the synergy of $\boldsymbol{b}$ and $\boldsymbol{\alpha}$, is confirmed by its strong monotonic correlation with the average output token length (Figure~\ref{fig:analysis_and_real_world}(d)). This intuitive result---as complex queries require elaborate responses---confirms our metric successfully captures task difficulty.

\subsubsection{Ablation Study on Anchor Sampling}
We analyze the effectiveness of different anchor sampling strategies, with results presented in Table 2. We observe that the Task-aware difficulty strategy achieves scores (0.32, -0.46, -0.44) comparable to the GraphRouter baseline. This validates our hypothesis that a comprehensive score combining both general difficulty ($\boldsymbol{b}$) and task-specific discrimination ($\boldsymbol{\alpha}$) is effective for model profiling. Moreover, the D-optimality strategy significantly outperforms all other variants, achieving the best scores (0.39, -0.34, -0.37). By explicitly maximizing Fisher information, our method selects the most diagnostically valuable samples, leading to the most efficient characterization of a new model's capabilities. Conversely, single-parameter strategies like Diff-based and Disc-based perform poorly, on par with Random Sampling (0.27, -0.56, -0.47).

\section{Conclusion}

We introduced ZeroRouter, a framework that successfully breaks model lock-in in LLM routing. By decoupling query properties from model capabilities via a universal latent space, our approach enables scalable, cost-efficient, and zero-shot integration of new models, making a new paradigm for adaptive LLM orchestration.

\section{Acknowledgments}
This work was supported by the National Natural Science Foundation of China (No. 62332016).

\bibliography{aaai2026}

@article{ding2024hybrid,
  title={Hybrid LLM: Cost-efficient and quality-aware query routing},
  author={Ding, Dujian and Mallick, Ankur and Wang, Chi and Sim, Robert and Mukherjee, Subhabrata and Ruhle, Victor and Lakshmanan, Laks VS and Awadallah, Ahmed Hassan},
  journal={arXiv preprint arXiv:2404.14618},
  year={2024}
}

@inproceedings{vsakota2024fly,
  title={Fly-swat or cannon? cost-effective language model choice via meta-modeling},
  author={{\v{S}}akota, Marija and Peyrard, Maxime and West, Robert},
  booktitle={Proceedings of the 17th ACM International Conference on Web Search and Data Mining},
  pages={606--615},
  year={2024}
}

@article{feng2024graphrouter,
  title={Graphrouter: A graph-based router for llm selections},
  author={Feng, Tao and Shen, Yanzhen and You, Jiaxuan},
  journal={arXiv preprint arXiv:2410.03834},
  year={2024}
}

@article{yao2025mmreason,
  title={MMReason: An Open-Ended Multi-Modal Multi-Step Reasoning Benchmark for MLLMs Toward AGI},
  author={Yao, Huanjin and Huang, Jiaxing and Qiu, Yawen and Chen, Michael K and Liu, Wenzheng and Zhang, Wei and Zeng, Wenjie and Zhang, Xikun and Zhang, Jingyi and Song, Yuxin and others},
  journal={arXiv preprint arXiv:2506.23563},
  year={2025}
}

@article{hendrycks2021measuring,
  title={Measuring mathematical problem solving with the math dataset},
  author={Hendrycks, Dan and Burns, Collin and Kadavath, Saurav and Arora, Akul and Basart, Steven and Tang, Eric and Song, Dawn and Steinhardt, Jacob},
  journal={arXiv preprint arXiv:2103.03874},
  year={2021}
}

@inproceedings{rein2024gpqa,
  title={Gpqa: A graduate-level google-proof q\&a benchmark},
  author={Rein, David and Hou, Betty Li and Stickland, Asa Cooper and Petty, Jackson and Pang, Richard Yuanzhe and Dirani, Julien and Michael, Julian and Bowman, Samuel R},
  booktitle={First Conference on Language Modeling},
  year={2024}
}

@article{li2024humaneval,
  title={HumanEval on Latest GPT Models--2024},
  author={Li, Daniel and Murr, Lincoln},
  journal={arXiv preprint arXiv:2402.14852},
  year={2024}
}

@article{wang2024mmlu,
  title={Mmlu-pro: A more robust and challenging multi-task language understanding benchmark},
  author={Wang, Yubo and Ma, Xueguang and Zhang, Ge and Ni, Yuansheng and Chandra, Abhranil and Guo, Shiguang and Ren, Weiming and Arulraj, Aaran and He, Xuan and Jiang, Ziyan and others},
  journal={Advances in Neural Information Processing Systems},
  volume={37},
  pages={95266--95290},
  year={2024}
}

@inproceedings{ong2024routellm,
  title={RouteLLM: Learning to Route LLMs from Preference Data},
  author={Ong, Isaac and Almahairi, Amjad and Wu, Vincent and Chiang, Wei-Lin and Wu, Tianhao and Gonzalez, Joseph E and Kadous, M Waleed and Stoica, Ion},
  booktitle={The Thirteenth International Conference on Learning Representations},
  year={2024}
}

@article{ni2024mixeval,
  title={MixEval: Deriving Wisdom of the Crowd from LLM Benchmark Mixtures},
  author={Ni, Jinjie and Xue, Fuzhao and Yue, Xiang and Deng, Yuntian and Shah, Mahir and Jain, Kabir and Neubig, Graham and You, Yang},
  journal={arXiv preprint arXiv:2406.06565},
  year={2024}
}

@article{polo2024tinybenchmarks,
  title={tinyBenchmarks: evaluating LLMs with fewer examples},
  author={Polo, Felipe Maia and Weber, Lucas and Choshen, Leshem and Sun, Yuekai and Xu, Gongjun and Yurochkin, Mikhail},
  journal={arXiv preprint arXiv:2402.14992},
  year={2024}
}

@article{suzgun2022challenging,
  title={Challenging big-bench tasks and whether chain-of-thought can solve them},
  author={Suzgun, Mirac and Scales, Nathan and Sch{\"a}rli, Nathanael and Gehrmann, Sebastian and Tay, Yi and Chung, Hyung Won and Chowdhery, Aakanksha and Le, Quoc V and Chi, Ed H and Zhou, Denny and others},
  journal={arXiv preprint arXiv:2210.09261},
  year={2022}
}

@article{zhou2023instruction,
  title={Instruction-following evaluation for large language models},
  author={Zhou, Jeffrey and Lu, Tianjian and Mishra, Swaroop and Brahma, Siddhartha and Basu, Sujoy and Luan, Yi and Zhou, Denny and Hou, Le},
  journal={arXiv preprint arXiv:2311.07911},
  year={2023}
}

@article{kipnis2024texttt,
  title={metabench--A Sparse Benchmark to Measure General Ability in Large Language Models},
  author={Kipnis, Alex and Voudouris, Konstantinos and Buschoff, Luca M Schulze and Schulz, Eric},
  journal={arXiv preprint arXiv:2407.12844},
  year={2024}
}

@inproceedings{rodriguez2021evaluation,
  title={Evaluation examples are not equally informative: How should that change NLP leaderboards?},
  author={Rodriguez, Pedro and Barrow, Joe and Hoyle, Alexander Miserlis and Lalor, John P and Jia, Robin and Boyd-Graber, Jordan},
  booktitle={Proceedings of the 59th Annual Meeting of the Association for Computational Linguistics and the 11th International Joint Conference on Natural Language Processing (Volume 1: Long Papers)},
  pages={4486--4503},
  year={2021}
}

@article{lin2021truthfulqa,
  title={Truthfulqa: Measuring how models mimic human falsehoods},
  author={Lin, Stephanie and Hilton, Jacob and Evans, Owain},
  journal={arXiv preprint arXiv:2109.07958},
  year={2021}
}

@article{clark2018think,
  title={Think you have solved question answering? try arc, the ai2 reasoning challenge},
  author={Clark, Peter and Cowhey, Isaac and Etzioni, Oren and Khot, Tushar and Sabharwal, Ashish and Schoenick, Carissa and Tafjord, Oyvind},
  journal={arXiv preprint arXiv:1803.05457},
  year={2018}
}

@article{natesan2016bayesian,
  title={Bayesian prior choice in IRT estimation using MCMC and variational Bayes},
  author={Natesan, Prathiba and Nandakumar, Ratna and Minka, Tom and Rubright, Jonathan D},
  journal={Frontiers in psychology},
  volume={7},
  pages={1422},
  year={2016},
  publisher={Frontiers}
}

@article{jiang2024mixtral,
  title={Mixtral of experts},
  author={Jiang, Albert Q and Sablayrolles, Alexandre and Roux, Antoine and Mensch, Arthur and Savary, Blanche and Bamford, Chris and Chaplot, Devendra Singh and Casas, Diego de las and Hanna, Emma Bou and Bressand, Florian and others},
  journal={arXiv preprint arXiv:2401.04088},
  year={2024}
}

@article{aggarwal2024automix,
  title={Automix: Automatically mixing language models},
  author={Aggarwal, Pranjal and Madaan, Aman and Anand, Ankit and Potharaju, Srividya Pranavi and Mishra, Swaroop and Zhou, Pei and Gupta, Aditya and Rajagopal, Dheeraj and Kappaganthu, Karthik and Yang, Yiming and others},
  journal={Advances in Neural Information Processing Systems},
  volume={37},
  pages={131000--131034},
  year={2024}
}

@article{mohammadshahi2024routoo,
  title={Routoo: Learning to Route to Large Language Models Effectively},
  author={Mohammadshahi, Alireza and Shaikh, Arshad Rafiq and Yazdani, Majid},
  journal={arXiv preprint arXiv:2401.13979},
  year={2024}
}

@article{cao2025should,
  title={How Should We Build A Benchmark? Revisiting 274 Code-Related Benchmarks For LLMs},
  author={Cao, Jialun and Chan, Yuk-Kit and Ling, Zixuan and Wang, Wenxuan and Li, Shuqing and Liu, Mingwei and Qiao, Ruixi and Han, Yuting and Wang, Chaozheng and Yu, Boxi and others},
  journal={arXiv preprint arXiv:2501.10711},
  year={2025}
}

@article{zhang2025capability,
  title={Capability instruction tuning: A new paradigm for dynamic llm routing},
  author={Zhang, Yi-Kai and Zhan, De-Chuan and Ye, Han-Jia},
  journal={arXiv preprint arXiv:2502.17282},
  year={2025}
}

@article{wang2025mixllm,
  title={Mixllm: Dynamic routing in mixed large language models},
  author={Wang, Xinyuan and Liu, Yanchi and Cheng, Wei and Zhao, Xujiang and Chen, Zhengzhang and Yu, Wenchao and Fu, Yanjie and Chen, Haifeng},
  journal={arXiv preprint arXiv:2502.18482},
  year={2025}
}

@misc{open-llm-leaderboard,
  author = {Edward Beeching and Clémentine Fourrier and Nathan Habib and Sheon Han and Nathan Lambert and Nazneen Rajani and Omar Sanseviero and Lewis Tunstall and Thomas Wolf},
  title = {Open LLM Leaderboard},
  year = {2023},
  publisher = {Hugging Face},
  howpublished = {\url{https://huggingface.co/spaces/HuggingFaceH4/open_llm_leaderboard}}
}

\end{document}